\documentclass[a4paper]{cas-sc}

\usepackage{float}  

\usepackage[numbers]{natbib}
\usepackage{booktabs}
\usepackage{graphicx}

\usepackage{algorithm}
\usepackage{algpseudocode}  
\usepackage{program}
\usepackage{tabularx}
\usepackage{makecell}
\usepackage{booktabs}
\usepackage{graphicx}
\usepackage{subcaption}
\usepackage{caption}
\usepackage{enumitem}
\usepackage{graphicx}  

\newlist{steps}{enumerate}{1}
\setlist[steps, 1]{label = Step \arabic*:}

\begin{document}

\shorttitle{Continual Learning in NLP and Catastrophic Forgetting}    


\title [mode = title]{A Comparative Empirical Study of Catastrophic Forgetting Mitigation in Sequential Task Adaptation for Continual Natural Language Processing Systems}  



%

\author[]{Aram Abrahamyan}



\ead{<aram_abrahamyan@edu.aua.am>}

\author[]{Sachin Kumar${^*}$}



\ead{<s.kumar@aua.am>}









\affiliation[]{organization={Zaven P. and Sonia Akian College of Science and Engineering},
         addressline={American University of Armenia}, 
          city={Yerevan},
          citysep={}, 
            postcode={0019}, 
           country={Armenia}}




\credit{equal}





\begin{abstract}
Neural language models deployed in real-world applications must continually adapt to new tasks and domains without forgetting previously acquired knowledge.
This work presents a comparative empirical study of catastrophic forgetting mitigation in continual intent classification.
Using the CLINC150 dataset, we construct a 10-task label-disjoint scenario and evaluate three backbone architectures: a feed-forward Artificial Neural Network (ANN), a Gated Recurrent Unit (GRU), and a Transformer encoder, under a range of continual learning (CL) strategies.
We consider one representative method from each major CL family: replay-based Maximally Interfered Retrieval (MIR), regularization-based Learning without Forgetting (LwF), and parameter-isolation via Hard Attention to Task (HAT), both individually and in all pairwise and triple combinations.
Performance is assessed with average accuracy, macro F1, and backward transfer, capturing the stability-plasticity trade-off across the task sequence.
Our results show that naive sequential fine-tuning suffers from severe forgetting for all architectures and that no single CL method fully prevents it.
Replay emerges as a key ingredient: MIR is the most reliable individual strategy, and combinations that include replay (MIR+HAT, MIR+LwF, MIR+LwF+HAT) consistently achieve high final performance with near-zero or mildly positive backward transfer.
The optimal configuration is architecture-dependent. MIR+HAT yields the best result for ANN and Transformer, MIR+LwF+HAT, on the other hand, works the best for GRU, and in several cases CL methods even surpass joint training, indicating a regularization effect.
These findings highlight the importance of jointly selecting backbone architecture and CL mechanism when designing continual intent-classification systems.
\end{abstract}




\maketitle
\section{Introduction and Background}

Neural language models are often trained on fixed datasets under the assumption that the task and data distribution remain stable. However, real-world NLP applications must operate in changing environments where new tasks and domains emerge over time. When models are updated with data from new tasks without revisiting earlier training examples, they tend to lose previously acquired knowledge. This phenomenon, known as \textit{catastrophic forgetting}, poses a major obstacle for building adaptive NLP systems.

Continual learning offers a framework for enabling models to learn from streams of tasks without forgetting earlier ones. Retraining on all past and present data is usually impractical due to computational cost, storage requirements, or privacy constraints, making continual learning a more viable solution. As reviewed in recent surveys \cite{r1}, continual learning methods are commonly grouped into three families: replay-based methods, regularization-based methods (RBM), and parameter-isolation methods (PIM).

Replay-based methods store a subset of samples from earlier tasks and interleave them with new data during training. This simple strategy can effectively preserve knowledge of previous tasks. More advanced approaches, such as Maximum Interfered Retrieval (MIR) \cite{r2}, select past samples based on their likelihood of interference with the current update. Lopez-Paz and Ranzato \cite{r3} demonstrate that replay-based approaches tend to outperform regularization-based techniques in many settings.

Regularization-based methods constrain changes to parameters that were important for earlier tasks. For example, Elastic Weight Consolidation (EWC) \cite{r4} penalizes large updates to parameters with high estimated importance, while Learning without Forgetting (LwF) \cite{r5} uses knowledge distillation to preserve the behaviour of previous models.

Parameter-isolation methods assign different subsets of model parameters to different tasks. This can be achieved by expanding the architecture for new tasks, as in Progressive Neural Networks~\cite{r6}, or by learning sparse task-specific masks that control neuron activations, as in PackNet~\cite{r7} and Hard Attention to Task (HAT)~\cite{r8}.

Although most early work in continual learning focused on computer vision, the field has gained increasing attention in NLP. Studies have explored replay, RBM, and PIM techniques for text classification, language modeling, and machine translation. However, less is known about how model architecture influences susceptibility to catastrophic forgetting and the effectiveness of different continual learning strategies.

Beyond these classical methods, a broader body of work has analyzed continual learning and its variants from both general and NLP-specific perspectives. Parisi et al. provide a comprehensive review of lifelong learning with neural networks \cite{r9}, while recent surveys discuss continual learning with pre-trained models and large-scale architectures \cite{r10}.
In NLP, several surveys highlight that continual learning must handle task, domain, and label shifts that differ from typical vision benchmarks \cite{r11,r12}. At the method level, episodic-memory and replay-based approaches such as MBPA++ \cite{r13} and language-model-based replay in LAMOL \cite{r14} have been proposed for lifelong language learning, and specialized regularization schemes have been developed for continual text classification \cite{r15}.

For intent classification and task-oriented dialogue systems, recent work has investigated continual, class-incremental, and open-world intent learning, including methods for incrementally adding new intents, handling out-of-scope queries, and performing generalized intent discovery \cite{r16, r17, r18, r19}. However, these studies rarely compare fundamentally different architectures under a unified continual-learning setup, leaving open how model design interacts with replay, regularization, and parameter-isolation strategies.

In this work, we study continual learning for intent classification using a controlled sequence of ten tasks with disjoint label sets. 

We compare three neural architectures that represent distinct modeling paradigms: a simple feed-forward Artificial Neural Network (ANN) using pooled embeddings, a recurrent Gated Recurrent Unit (GRU) model that captures temporal dependencies, a Transformer encoder based on self-attention. Together, these models span a spectrum from lightweight feed-forward computation to recurrence and global attention.

The main objective of this study is to find answers to the following reseach questions:
\begin{itemize}
  \item How do continual learning (CL) methods compare to naive sequential fine-tuning and joint training?
  \item How do replay, regularization, and parameter-isolation-based strategies differ in terms of stability and plasticity?
  \item To what extent does the underlying architecture influence continual learning behaviour?
\end{itemize}

To address these questions, we adopt the following experimental design:
\begin{enumerate}
    \item Construct ten label-disjoint tasks from the benchmark intent-classification dataset and define consistent train/validation/test splits for each task.
    \item Apply a shared preprocessing and tokenization pipeline and prepare data loaders for all tasks.
    \item For each architecture (ANN, GRU, Transformer), tune global hyperparameters and train baseline models under joint training and naive sequential fine-tuning.
    \item Train continual learning models using replay (MIR), regularization-based (LwF), parameter-isolation (HAT) methods, and their pairwise combinations across all ten tasks.
    \item Record the performance matrix $R_{i,j}$ for accuracy and macro F1, and compute AA, AF1, BWT, FWT, and forgetting measures.
    \item Analyze how stability and plasticity vary across methods and architectures, and identify which combinations yield the best trade-offs for continual intent classification.
\end{enumerate}

\section{Dataset Description}

The experiments in this study use the CLINC150 dataset~\cite{r20}, a publicly available benchmark designed for intent detection in task-oriented dialogue systems. The dataset is released in four configurations, and for this work we use the full version, which contains 23,700 utterances grouped into 150 distinct intents. These intents span ten broader domains, making the dataset both diverse and representative of real conversational scenarios. Each entry consists of a short user utterance paired with an intent label, and the data is provided with predefined training, validation, and test splits containing 15,000, 3,000, and 4,500 samples respectively. 

In addition to in-scope utterances, CLINC150 includes a set of out-of-scope examples intended to assess whether models can recognize inputs that fall outside the known intent space. Since the focus of this project is continual learning over clearly defined intents, these out-of-scope samples are removed. They account for 100 training examples, 100 validation examples, and 1,000 test examples. After discarding them, the remaining dataset is well balanced, with each intent represented by a similar number of samples. This balance is beneficial for training models without introducing prior bias toward particular labels.

The dataset is cleanly formatted in JSON and contains no missing values, so only modest preprocessing is required. All text is lowercased to maintain consistency, and simple cleaning operations are applied to remove URLs and unnecessary whitespace while keeping the text suitable for modern transformer tokenizers. The user utterances are then processed using the Hugging Face DistilBERT tokenizer, which produces token indices and attention masks. Intent labels are converted into numerical identifiers through label encoding, and all token sequences are padded or truncated to a fixed length of 64 tokens to ensure uniform input dimensions for the neural models.

An exploratory analysis of the dataset confirms that, aside from the out-of-scope category that is excluded from further experimentation, the distribution of samples across the 150 intents is highly uniform. This balanced structure supports fair evaluation of continual learning methods, since differences in performance can be attributed to the methods themselves rather than to imbalance in the underlying data.

Moreover, we found out that the average length of user utterances is around 40 words with the minimum being 2 words and maximum being 136 words long, with some variation across different intents.
This indicates that the dataset contains a mix of short and moderately long queries.
The median of the utterance length is 38 words which itself indicates that the distribution of the lengths of utternaces is skewed just a little bit, it is more or less bell-shaped.
The half of the samples have lengths between 27 and 52 words, that is why we chose the maximum sequence length to be 64.
We found out that the dataset has a rich vocabulary, with many unique words used across different intents (total of 7446 unique words). This diversity can help in training models that generalize well to unseen data.
See Figure \ref{fig: Utterance Length Distribution in CLINC150 Dataset} for more details about the distribution of utterance lengths.
\begin{center}
  \includegraphics[scale=0.5]{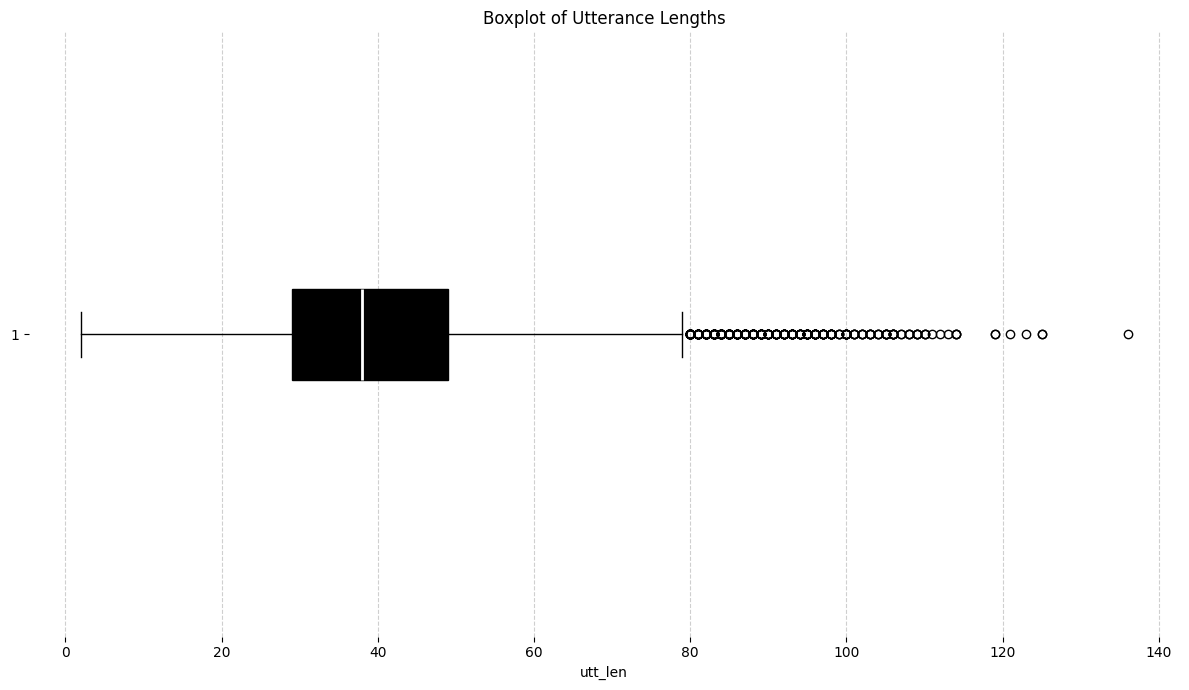}
  \captionof{figure}{Utterance Length Distribution in CLINC150 Dataset}
  \label{fig: Utterance Length Distribution in CLINC150 Dataset}
\end{center}

Furthermore, there are a lot of stopwords in the dataset (100754 out of 202057 words are stopwords), which we decided not to remove to keep input distribution as close as possible to the real-world data distribution and to what moder neural models such as Transformers are trained on.
Many so-called stopwords, for example, auxiliaries, negations, prepositions carry important intent and semantic information, so removing could distort meaning of the sentences, whihc is undesirable. 
Figure \ref{fig: 20 Most Frequent Words in CLINC150 Dataset} shows that the most frequent words along all the samples are stopwords.

\begin{center}
    \includegraphics[scale=0.15]{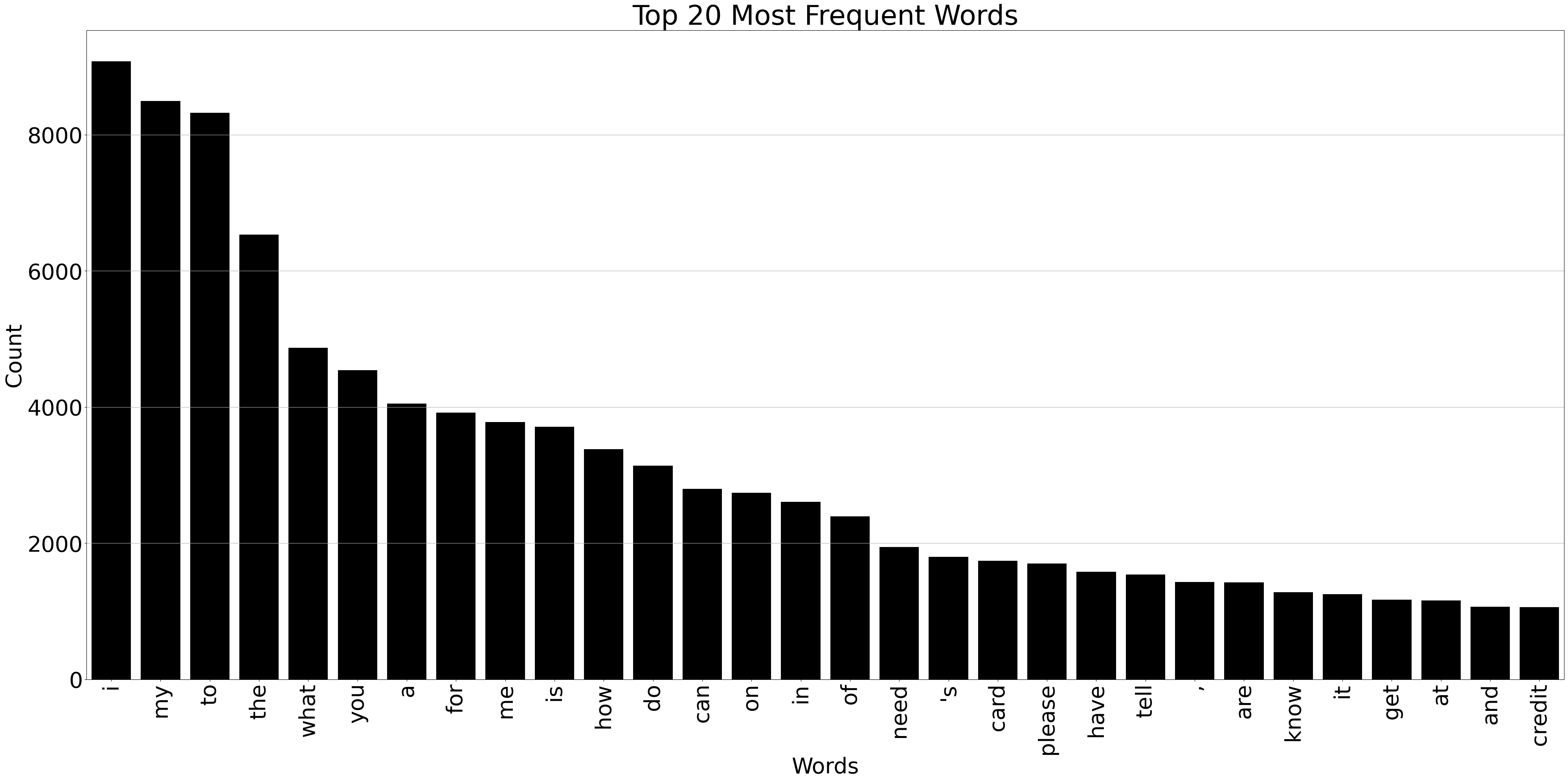}
    \captionof{figure}{20 Most Frequent Words in CLINC150 Dataset}
    \label{fig: 20 Most Frequent Words in CLINC150 Dataset}
  \end{center}

\section{Methodology and Research Plan}
\label{sec:methodology}
In this section we describe the continual learning setting, model architectures, continual learning methods, and the overall research plan adopted in this work.
\subsection{Continual Learning Setting}
We consider a supervised continual learning scenario with $T=10$ sequential tasks $\{\mathcal{T}_t\}_{t=1}^{10}$. As described in the Dataset section, the full set of intent labels is partitioned into 10 disjoint subsets, each defining one task with its own train, validation, and test split. Formally, if $\mathcal{Y}$ is the set of all labels, then
\begin{equation*}
  \mathcal{Y} = \bigcup_{t=1}^{10} \mathcal{Y}_t, \qquad \mathcal{Y}_i \cap \mathcal{Y}_j = \emptyset \ \text{for } i \neq j,
\end{equation*}
and task $\mathcal{T}_t$ is a standard multi-class classification problem over labels $\mathcal{Y}_t$.
The model is trained on the tasks in order $\mathcal{T}_1,\dots,\mathcal{T}_{10}$.
After finishing training on task $t$, we evaluate the current model on all test sets $\mathcal{T}_1,\dots,\mathcal{T}_t$. This allows us to track both performance on the current task and potential forgetting on earlier tasks.

\begin{algorithm}[h]
  \caption{Task construction on CLINC150}
  \label{alg:task_construction}
  \begin{algorithmic}[1]
  \Require CLINC150 dataset $\mathcal{D}$ with label set $\mathcal{Y}$
  \Require Number of tasks $T = 10$
  
  \State Partition $\mathcal{Y}$ into disjoint subsets $\{\mathcal{Y}_t\}_{t=1}^T$
  \Statex \hspace{1.5em} with $\mathcal{Y}_i \cap \mathcal{Y}_j = \emptyset$ for $i \neq j$
  \For{$t = 1$ to $T$}
    \State Build task $\mathcal{T}_t = (\mathcal{D}_t^{\text{train}}, \mathcal{D}_t^{\text{val}}, \mathcal{D}_t^{\text{test}})$
    \Statex \hspace{1.5em} using only labels in $\mathcal{Y}_t$
  \EndFor
  \end{algorithmic}
  \end{algorithm}

\subsection{Preprocessing and Representation}
The preprocessing is kept minimal to retain the natural text distribution.
We lowercase all text, remove URLs, and collapse multiple spaces.
We do not perform stopword removal or stemming/lemmatization since many function words and negations carry important intent information and we aim to avoid unnecessary distribution shifts in a continual learning setting.
Tokenization is performed using the AutoTokenizer from Hugging Face's Transformers library (DistilBERT) to produce \textsc{input\_ids} and \textsc{attention\_mask}.
Intent labels are converted to numerical format using label encoding.
Sequences are padded or truncated to a fixed length of 64 tokens (50\% of the utterances have length in range of 27 to 52).

\begin{algorithm}[h]
  \caption{Preprocessing and tokenization}
  \label{alg:preproc}
  \begin{algorithmic}[1]
  \Require Tasks $\{\mathcal{T}_t\}_{t=1}^T$, max length $L$
  
  \For{each utterance $x$ in all $\mathcal{D}_t$}
    \State Lowercase $x$, remove URLs, collapse multiple whitespace
    \State Tokenize $x$ with a DistilBERT tokenizer
    \State Pad or truncate tokens to fixed length $L$ (e.g., $L = 64$)
  \EndFor
  \State Encode all intent labels $y$ as integer IDs shared across tasks
  \end{algorithmic}
\end{algorithm}

\subsection{Model Architectures}
We compare three neural architectures that represent distinct modelling paradigms for text (\textit{See Figure \ref{fig:archs}}).
First, we use a simple feed-forward Artificial Neural Network (ANN as a lightweight baseline with limited sequence-modelling capacity, to reveal how a non-sequential model behaves under continual learning.
Second, we employ a Gated Recurrent Unit (GRU) network, which explicitly encodes word order and local temporal dependencies and is representative of classic recurrent sequence models widely used in earlier NLP systems.
Third, we consider a Transformer encoder, which models long-range and global token interactions via self-attention and reflects modern attention-based architectures for text classification.
We select ANN, GRU, and Transformer because they span three complementary modelling paradigms: shallow feed-forward, recurrent, and attention-based. This spectrum enables us to investigate whether continual learning performance is driven primarily by the CL strategy or is also strongly shaped by the underlying architectural inductive biases and capacity.

\begin{figure}[ht]
  \centering
  \setlength{\fboxsep}{6pt}  
  \setlength{\fboxrule}{0.8pt} 
  
  \fbox{\parbox{1.5cm}{\centering Token IDs}}
  \,$\longrightarrow$\,
  \fbox{\parbox{1.5cm}{\centering Embedding\\Layer}}
  \,$\longrightarrow$\,
  \fbox{\parbox{1.5cm}{\centering Mean\\Pooling}}
  \,$\longrightarrow$\,
  \fbox{\parbox{1.5cm}{\centering \textbf{Fully Connected}\\\textbf{Hidden Layer}}}
  \,$\longrightarrow$\,
  \fbox{\parbox{1.5cm}{\centering ReLU Activation\\+\\Dropout}}
  \,$\longrightarrow$\,
  \fbox{\parbox{1.5cm}{\centering Output\\Layer}}

\vskip 1em

  \fbox{\parbox{1.5cm}{\centering Token IDs}}
  \,$\longrightarrow$\,
  \fbox{\parbox{1.5cm}{\centering Embedding\\Layer}}
  \,$\longrightarrow$\,
  \fbox{\parbox{1.5cm}{\centering \textbf{GRU Layer}}}
  \,$\longrightarrow$\,
  \fbox{\parbox{1.5cm}{\centering Output\\Layer}}

  \vskip 1em
  
  \fbox{\parbox{1.5cm}{\centering Token IDs}}
  \,$\longrightarrow$\,
  \fbox{\parbox{1.5cm}{\centering Embedding\\Layer}}
  \,$\longrightarrow$\,
  \fbox{\parbox{1.5cm}{\centering\textbf{Transformer}\\\textbf{Encoder}}}
  \,$\longrightarrow$\,
  \fbox{\parbox{1.5cm}{\centering Mean\\Pooling}}
  \,$\longrightarrow$\,
  \fbox{\parbox{1.5cm}{\centering Output\\Layer}}
  
  \caption{Schematic diagram of ANN, GRU, and Transformer encoder}
  \label{fig:archs}
  \end{figure} 
\newpage
\subsection{Training Protocol and Evaluation Metrics}
\label{sec:methodology-metrics}
For each architecture (ANN, GRU, Transformer) we train:
\begin{itemize}
  \item \textbf{Baselines:}
  \begin{itemize}
      \item \emph{Joint training}, where a single model is trained on the union of all tasks. This serves as an upper bound without continual constraints.
      \item \emph{Naive sequential fine-tuning}, where tasks are learned one after another with no CL mechanism.
  \end{itemize}
  \item \textbf{Continual learning runs:} Replay (MIR), RBM (LwF), PIM (HAT), and all combinations mentioned above.
\end{itemize}
Global hyperparameters (learning rate, batch size, number of epochs, patience, etc.) are tuned using Bayesian optimization with a fixed seed of 42 on validation splits from the early tasks, and then fixed for the remaining experiments for fair comparison.
The hardware used includes an NVIDIA RTX 3070 GPU, 64GB RAM, and an Intel i7 11 Gen CPU.
Let $R_{i,j}$ denote the performance on task $j$ after finishing training on task $i$. We use the following metrics:

\begin{itemize}
  \item \textbf{Average Accuracy (AA)} after the final task\\
AA measures the overall classification performance of a continual learning model after it has seen all $T$ tasks.
The term $R_{T,j}$ denotes the test accuracy on task $j$ after training has finished on the final task $T$, and the metric simply averages this value over all tasks.
A high AA indicates that the model retains good performance across the entire task sequence at the end of training.
In this work, AA serves as a global summary of how well each architecture--strategy combination preserves task performance after sequential training.

  \[
  \mathrm{AA} = \frac{1}{T} \sum_{j=1}^{T} R_{T,j}.
  \]
  \item \textbf{Average F1 Score (AF1)} after the final task\\
  AF1 is defined analogously to AA, but uses the (macro-)F1 score $F1_{T,j}$ instead of accuracy.
  Because F1 accounts for both precision and recall and is less sensitive to class imbalance than accuracy, it provides a more informative measure in multi-class intent classification, where some intents are less frequent in case of some CL methods.
  In our experiments, AF1 complements AA by highlighting cases where a method may achieve good accuracy but still perform poorly on minority classes.
  \[
  \mathrm{AF1} = \frac{1}{T} \sum_{j=1}^{T} F1_{T,j}.
  \]
  \item \textbf{Backward Transfer with Accuracy or F1-Score (BWT)}\\
BWT quantifies how learning new tasks influences performance on previously learned tasks.
  For each task $j < T$, we compare the final performance on task $j$ after completing all tasks, $R^{(\text{acc/F1})}_{T,j}$, with the performance on $j$ immediately after it was first learned, $R^{(\text{acc/F1})}_{j,j}$.
  Positive BWT indicates beneficial backward transfer (improvements on past tasks due to later training), whereas negative BWT reflects degradation, i.e., catastrophic forgetting.
  In this work, we use BWT (computed with both accuracy and F1) to directly assess the degree of forgetting or positive transfer induced by each continual learning strategy.
  \[
  \mathrm{BWT_{acc/F1}} = \frac{1}{T-1} \sum_{j=1}^{T-1} \big( R^{(acc/F1)}_{T,j} - R^{(acc/F1)}_{j,j} \big).
  \]
  \item \textbf{Final Average Backward Transfer $(\overline{BWT})$}\\
  While BWT is defined per task, $\overline{\text{BWT}}_{\text{acc/F1}}$ aggregates these values over the whole sequence to yield a single scalar that summarizes the overall backward transfer behaviour of a model.
  Here, $\text{BWT}_i$ denotes the average backward transfer up to task $i$, and $\overline{\text{BWT}}_{\text{acc/F1}}$ averages these quantities over all $T$ tasks.
  This metric is particularly useful in our study because it allows a compact comparison of architectures and strategies in terms of their overall tendency to forget or improve past tasks throughout training, rather than only at individual time points.
  \[
  \overline{\mathrm{BWT_{acc/F1}}} = \frac{1}{T} \sum_{i=1}^{T} \mathrm{BWT_{acc/F1}}_i, \quad \text{where } \mathrm{BWT}_i = \frac{1}{i-1} \sum_{j=1}^{i-1} \big( R^{(acc/F1)}_{i,j} - R^{(acc/F1)}_{j,j} \big).
  \]
\end{itemize}
These metrics allow us to quantify the trade-off between \emph{stability} (retaining old tasks) and \emph{plasticity} (learning new tasks).

\begin{algorithm}[h]
  \caption{Metric computation for pair $(a,c)$}
  \label{alg:metrics}
  \begin{algorithmic}[1]
  \Require $R^{(\text{acc})}$, $R^{(\text{F1})}$
  
  \State $\mathrm{AA} = \dfrac{1}{T} \sum_{j=1}^{T} R^{(\text{acc})}_{T,j}$
  \State $\mathrm{AF1} = \dfrac{1}{T} \sum_{j=1}^{T} R^{(\text{F1})}_{T,j}$
  \State $\overline{\mathrm{BWT}}_{\text{acc}} = \dfrac{1}{T-1} \sum_{j=1}^{T-1}
          \big( R^{(\text{acc})}_{T,j} - R^{(\text{acc})}_{j,j} \big)$
  \State Analogously compute $\overline{\mathrm{BWT}}_{\text{F1}}$
  \end{algorithmic}
  \end{algorithm}

\subsection{Continual Learning Methods}
To mitigate catastrophic forgetting, we implement and evaluate three representative continual learning methods from each family described in the Literature Review section, namely Maximally Interfered Retrieval (MIR), Learning without Forgetting (LwF), and Hard Attention to Task (HAT).
\subsubsection{Replay: Maximally Interfered Retrieval (MIR)}
The replay-based method maintains a memory buffer of samples from previous tasks.
However, those samples are not selected randomly but the most hurt by learning the new task.
For the first step of learning a new task, we do a virtual update on the current batch $S$, meaning we pretend we take one gradient step:
\begin{equation*}
  \theta' = \theta - \alpha \nabla_\theta \mathcal{L}(f_\theta, S)
\end{equation*}
Then, we measure how much each memory sample is hurt.
For each stored (old) example $(x_i, y_i)$ in the memory buffer, we compute the loss before and after the virtual update:
\begin{equation*}
  \mathcal{L}(x_i, y_i, \theta) \enskip \& \enskip \mathcal{L}(x_i, y_i, \theta')
\end{equation*}
Define $r_i = \mathcal{L}(x_i, y_i, \theta) - \mathcal{L}(x_i, y_i, \theta')$.
If $r_i > 0$, it means this update made the old example worse and we encountered catastrophic forgetting.
Then we select the top $k$ samples with the highest $r_i$ values (the most forgotten samples) to replay alongside the new task data.
The samples with the highest interference are the most informative to rehearse to mitigate forgetting.
\begin{algorithm}[h]
  \footnotesize
  \caption{MIR: targeted replay subroutine}
  \label{alg:mir}
  \begin{algorithmic}[1]
  \Function{MIRReplay}{$S_t, B, c, \theta$}
    \If{$c$ uses MIR and $B$ is not empty}
      \State $\theta' \gets \theta - \eta \nabla_{\theta} \mathcal{L}_{\text{task}}(S_t; \theta)$
      \For{each $(x_m, y_m) \in B$}
        \State $r_m \gets \mathcal{L}_{\text{task}}(x_m, y_m; \theta') - \mathcal{L}_{\text{task}}(x_m, y_m; \theta)$
      \EndFor
      \State $M_t \gets \text{TopK}(\{(x_m, y_m)\}, \{r_m\})$
      \State $S_t \gets S_t \cup M_t$
    \EndIf
    \State \Return $S_t$
  \EndFunction
  \end{algorithmic}
  \end{algorithm}

\subsubsection{Regularization-Based Method (RBM): Learning without Forgetting (LwF)}
The regularization-based method keeps a copy of the model before training on a new task (the ``teacher'') with parameters $\theta_{old}$ and trains a new model (the ``student'') with parameters $\theta_{new}$ on the new task data.
For an input $x$, let $z_{\text{old}}(x)$ and $z_{\text{new}}(x)$ denote the teacher and student logits.
Using a temperature $T>1$, we define:
\[
p^T_{old}(x) = softmax\left(\frac{z_{old}(x)}{T}\right), \quad p^T_{new}(x) = softmax\left(\frac{z_{new}(x)}{T}\right)
\]
The total loss for training on the new task is a combination of the standard cross-entropy loss on the new task labels and a Kullback--Leibler divergence term:
\[
\mathcal{L} = \mathcal{L}_{\text{task}} + \alpha T^2 \cdot\, \mathrm{KL}\big(p_{\text{old}}^{T}(x) \,\|\, p_{\text{new}}^{T}(x)\big),
\]
where $\alpha$ controls the strength of the regularization. This Learning without Forgetting (LwF) objective penalizes updates that would drastically alter the output distribution learned on previous tasks, even though no raw data from those tasks is stored.

\begin{algorithm}[h]
  \caption{LwF: distillation loss subroutine}
  \label{alg:lwf}
  \begin{algorithmic}[1]
  \Function{LwFLoss}{$\mathcal{L}_{\text{task}}, S_t, z, \theta^{\text{old}}, c$}
    \If{$c$ uses LwF}
      \For{each $(x,y)\in S_t$}
        \State Compute teacher logits $z_{\text{old}}(x)$ from $\theta^{\text{old}}$
        \State $p_{\text{old}}^{T}(x) = \mathrm{softmax}(z_{\text{old}}(x) / T)$
        \State $p_{\text{new}}^{T}(x) = \mathrm{softmax}(z(x) / T)$
      \EndFor
      \State $\mathcal{L}_{\text{dist}} \gets \frac{1}{\lvert S_t\rvert} \sum_{(x,y)\in S_t} T^2
              \, \mathrm{KL}(p_{\text{old}}^{T}(x) \,\|\, p_{\text{new}}^{T}(x))$
      \State $\mathcal{L} \gets \mathcal{L}_{\text{task}} + \alpha \mathcal{L}_{\text{dist}}$
    \Else
      \State $\mathcal{L} \gets \mathcal{L}_{\text{task}}$
    \EndIf
    \State \Return $\mathcal{L}$
  \EndFunction
  \end{algorithmic}
  \end{algorithm}

\subsubsection{Parameter Isolation Method (PIM): Hard Attention to Task (HAT)}
The parameter isolation method uses Hard Attention to Task (HAT), which learns task-specific masks that gate neuron activations.
For each layer $l$ and task $t$, a task embedding is transformed into a gate vector $a^{(l,t)} = \sigma(s\cdot e^{(l,t)})$, where $e^{(l,t)}$ is the task embedding, $\sigma$ is the sigmoid function, and $s$ is a scaling hyperparameter.
Activations are masked as:
\[
h'^{(l)} = a^{(l,t)} \odot h^{(l)},
\]
where $\odot$ denotes element-wise multiplication, $h^{(l)}$ are the original activations at layer $l$.
After training task $t$, we maintain cumulative masks that record which units have been heavily used so far.
\[
m^{(l)}_{\leq t} = \max(m^{(l)}_{\leq t -1}, a^{(l,t)})
\]
$m^{(l)}_{\leq t}$ shows how much each neuron in layer $l$ has been used by tasks up to $t$.
During training of task $t+1$ gradients are scaled down where $m^{(l)_{\leq t}}$ is high.
Thus, those neurons belonging to previous tasks are protected from significant updates, effectively isolating parameters for different tasks.

\begin{algorithm}[h]
  \caption{HAT: task-specific masking subroutine}
  \label{alg:hat}
  \begin{algorithmic}[1]
  \Function{HATMasking}{$\mathcal{L}, c$}
    \If{$c$ uses HAT}
      \For{each layer $l$}
        \State $a^{(l,t)} = \sigma(s \cdot e^{(l,t)})$
        \State Mask activations $h^{(l)} \gets a^{(l,t)} \odot h^{(l)}$
        \State During backprop, scale gradients using cumulative masks
      \EndFor
    \EndIf
    \State \Return $\mathcal{L}$  \Comment{unchanged in form, but gradients are masked}
  \EndFunction
  \end{algorithmic}
  \end{algorithm}
\newpage
\subsubsection{Combined Strategies}
In addition to evaluating MIR, LwF, and HAT individually, we also consider their combinations (Replay+RBM, Replay+PIM, RBM+PIM, and Replay+RBM+PIM). In these settings, memory replay, distillation, and masking are applied jointly. This enables us to quantify whether hybrid methods offer better stability--plasticity trade-offs than any single strategy.
\begin{algorithm}[h]
  \small
  \caption{Continual learning training loop}
  \label{alg:cl_main}
  \begin{algorithmic}[1]
  \Require Architectures $\mathcal{A}$, CL methods $\mathcal{C}$  
  \Require Tasks $\{\mathcal{T}_t\}_{t=1}^T$

  \For{each architecture $a \in \mathcal{A}$}
    \State Fix hyperparameters $\phi_a^\star$
    \For{each CL method $c \in \mathcal{C}$}
      \State Initialize model parameters $\theta$ for $a$
      \State Initialize CL-specific state (buffer $B$, teacher $\theta^{\text{old}}$, HAT masks)

      \For{$t = 1$ to $T$} \Comment{sequential tasks}
        \While{not early-stopped on $\mathcal{D}_t^{\text{val}}$}

          \If{$c$ uses MIR}
          \State Sample minibatch $S_t$ from $\mathcal{D}_t^{\text{train}}$
            \State $S_t \gets \textsc{MIRReplay}(S_t, B, \theta)$
          \EndIf

          \State $(\mathcal{L}_{\text{task}}, z) \gets \textsc{ForwardAndTaskLoss}(S_t, \theta)$
          \State $\mathcal{L} \gets \mathcal{L}_{\text{task}}$

          \If{$c$ uses LwF}
            \State $\mathcal{L} \gets \textsc{LwFLoss}(\mathcal{L}, S_t, z, \theta^{\text{old}})$
          \EndIf

          \If{$c$ uses HAT}
            \State $\mathcal{L} \gets \textsc{HATMasking}(\mathcal{L}, t)$
          \EndIf

          \State Update parameters $\theta \gets \theta - \eta \nabla_\theta \mathcal{L}$

          \If{$c$ uses MIR}
            \State Update memory buffer $B$ with samples from $S_t$
          \EndIf
        \EndWhile

        \State Evaluate on tasks $1..t$ and fill $R^{(\text{acc})}, R^{(\text{F1})}$
        \If{$c$ uses LwF}
          \State $\theta^{\text{old}} \gets \theta$
        \EndIf
      \EndFor

      \State Compute AA, AF1, $\overline{\mathrm{BWT}}$ using Algorithm~\ref{alg:metrics}
    \EndFor
  \EndFor
  \end{algorithmic}
\end{algorithm}
\newpage
\section{Experiments and Results}
In this section we describe the experimental setup and present quantitative results for the three architectures (ANN, GRU, Transformer) combined with the continual learning methods introduced in Section~\ref{sec:methodology}.
All experiments are implemented in Python using PyTorch and Hugging Face's Transformers library.
All models are trained on the 10-task split of the CLINC150 dataset described above.
For each architecture (ANN, GRU, Transformer), we first tune global hyperparameters (learning rate, batch size, number of epochs, patience for early stopping) using Bayesian optimization with a fixed seed of 42 on the full dataset.
These hyperparameters are then fixed for all subsequent experiments to ensure fair comparison across methods.
For each task, we use the provided training set and create a validation split by holding out a fixed proportion of the training examples.
Training proceeds task by task: after finishing task $t$, we evaluate the model on the test sets of all tasks $1,\dots,t$ and store the resulting accuracy and macro F1 scores in the performance matrix $R_{i,j}$.
From this matrix we compute Average Accuracy (AA), Backward Transfer (BWT), and Average F1-score (AF1) as defined in Section~\ref{sec:methodology-metrics}.

We begin by comparing two baselines (Joint training and Naive sequential fine-tuning) described in Section~\ref{sec:methodology-metrics} for each architecture.
\begin{center}
  \captionof{table}{Baseline Results on CLINC150 Dataset}
  \label{tab:baseline_results}
  \begin{tabular}{p{3cm}p{5cm}p{5cm}}
      \toprule
      \textbf{Model} & \textbf{Joint Training} & \textbf{Naive Sequential Fine-Tuning} \\
      \midrule
      Artificial Neural Network (ANN) & $Accuracy = 0.877$ & $AA = 0.1004, \enskip AF1 = 0.104$ \\
      \midrule
      Gated Recurrent Unit (GRU) & $Accuracy = 0.845$ & $AA = 0.0916,\enskip AF1=0.0915$\\
      \midrule
      Transformer Encoder & $Accuracy = 0.8707$ & $AA = 0.1018, \enskip AF1 = 0.1048$\\
      \bottomrule
  \end{tabular}
\end{center}
As expected, joint training achieves the highest overall performance, while naive sequential fine-tuning suffers from substantial catastrophic forgetting.
We did not include BWT for naive fine-tuning since it is overwhelmingly negative due to forgetting.
We can Interpret the poorer performance of Transformer compared to ANN in joint training as Transformer is a data-hungry model and may require more data or pretraining to fully leverage its capacity.
See Figure~\ref{fig:naive_acc} for details.
\begin{center}
  \includegraphics[width=\textwidth]{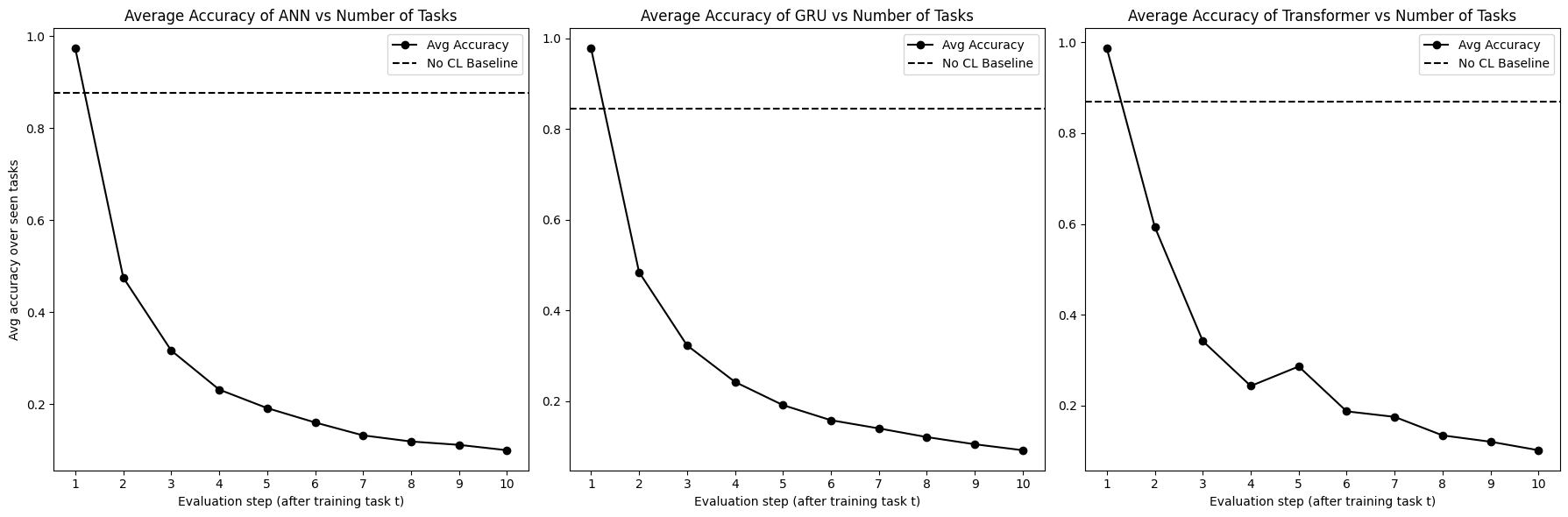}
  \captionof{figure}{Avarage Accuracy change over tasks for Naive Sequential training for each architecture.}
  \label{fig:naive_acc}
\end{center}
The next set of experiments evaluates the three continual learning methods (MIR, LwF, HAT) individually across all three architectures.
\newpage
\begin{center}
  \captionof{table}{MIR, LwF, and HAT Results on CLINC150 Dataset}
  \label{tab:indiv_results}
  \begin{tabular}{p{1.5cm}p{4.6cm}p{4.6cm}p{4.6cm}}
      \toprule
      \textbf{Model} & \textbf{MIR} & \textbf{LwF} & \textbf{HAT} \\
      \midrule
      Artificial Neural Network (ANN) &
      \makecell[l]{$AA = 0.193,\enskip AF1=0.227,$\\ $\overline{BWT_{acc}} = -0.7678$\\ $\overline{BWT_{F1}} = -0.7243$} &
      \makecell[l]{$AA = 0.0071,\enskip AF1=0.0017,$\\ $\overline{BWT_{acc}} = -0.3415$\\ $\overline{BWT_{F1}} = -0.3476$} &
      \makecell[l]{$AA = 0.4133,\enskip AF1=0.4507,$\\ $\overline{BWT_{acc}} = -0.4991$\\ $\overline{BWT_{F1}} = -0.4470$}\\
      \midrule
      Gated Recurrent Unit (GRU) &
      \makecell[l]{$AA = 0.5849,\enskip AF1=0.6906,$\\ $\overline{BWT_{acc}} = -0.2271$\\ $\overline{BWT_{F1}} = -0.1444$}
      & \makecell[l]{$AA = 0.1889,\enskip AF1=0.2191,$\\ $\overline{BWT_{acc}} = -0.5568$\\ $\overline{BWT_{F1}} = -0.5399$}
      & \makecell[l]{$AA = 0.0942,\enskip AF1=0.0962,$\\ $\overline{BWT_{acc}} = -0.9573$\\ $\overline{BWT_{F1}} = -0.9544$}\\
      \midrule
      Transformer Encoder
      & \makecell[l]{$AA = 0.7444,\enskip AF1=0.8217,$\\ $\overline{BWT_{acc}} = -0.1592$\\ $\overline{BWT_{F1}} = -0.1017$}
      & \makecell[l]{$AA = 0.5253,\enskip AF1=0.6218,$\\ $\overline{BWT_{acc}} = -0.1334$\\ $\overline{BWT_{F1}} = -0.1168$}
      & \makecell[l]{$AA = 0.1302,\enskip AF1=0.1419,$\\ $\overline{BWT_{acc}} = -0.906$\\ $\overline{BWT_{F1}} = -0.8849$}\\
      \bottomrule
  \end{tabular}
\end{center}

For the ANN, HAT clearly outperforms the other methods, reaching 
$AA=0.41$ and $AF1=0.45$, while MIR remains much weaker and LwF almost completely collapses (near-zero accuracy and F1).
However, even with HAT the ANN still shows strong forgetting, with $\overline{BWT}\approx -0.5.$

For the GRU, the best results are obtained with MIR ($AA\approx 0.58$, $AF1\approx 0.69$), with substantially milder forgetting ($BWT_{acc}\approx-0.23$, $BWT_{F1}\approx -0.14$) compared to LwF and especially HAT, which both degrade heavily.
This suggests that replay is particularly well suited to recurrent models in this setting.

For the Transformer, MIR again achieves the strongest performance, with $AA\approx 0.74$ and $AF1\approx 0.82$ and relatively small negative backward transfer. LwF provides moderate gains($AA\approx0.53$, $AF1\approx0.62$) with slightly improved BWT, whereas HAT fails to learn the task sequence effectively.
Overall, MIR is the most reliable method across architectures, HAT only helps the simple ANN, and LwF yields intermediate improvements, especially for the Transformer, while none of the methods fully eliminates forgetting.
However, in some cases we observe that though the average accuracy is low, the backward transfer is close to 0, meaning that the model is stable but not plastic enough to learn new tasks (e.g. GRU+MIR, Transformer + MIR, Transformer+LwF).
\begin{center}
  \includegraphics[width=\textwidth]{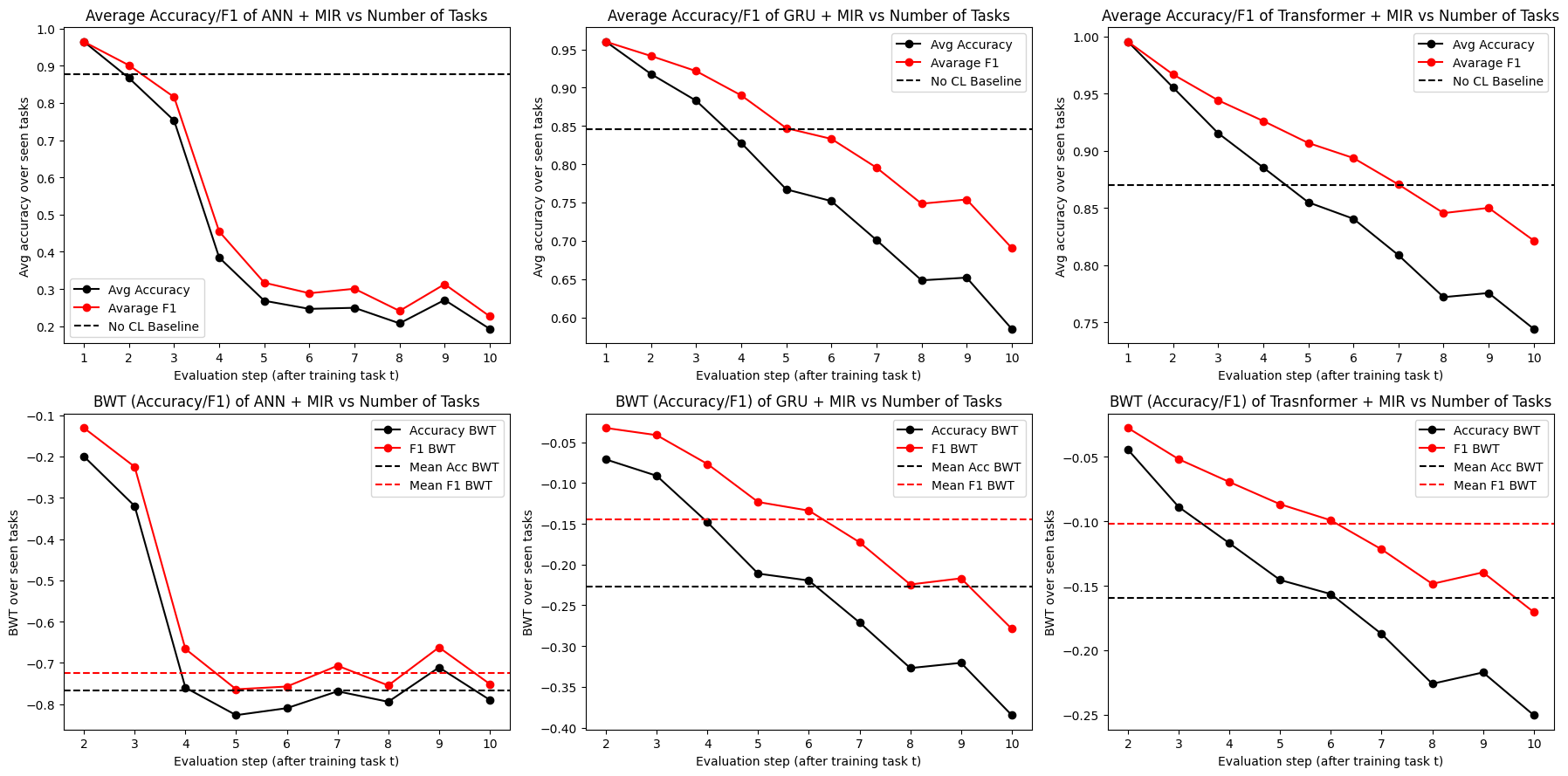}
  \captionof{figure}{AA, AF1, and average BWT over accuracy and F1 as they change over tasks for sequential training with MIR across all architectures.}
  \label{fig:mir_plot}
\end{center}

\begin{center}
  \includegraphics[width=\textwidth]{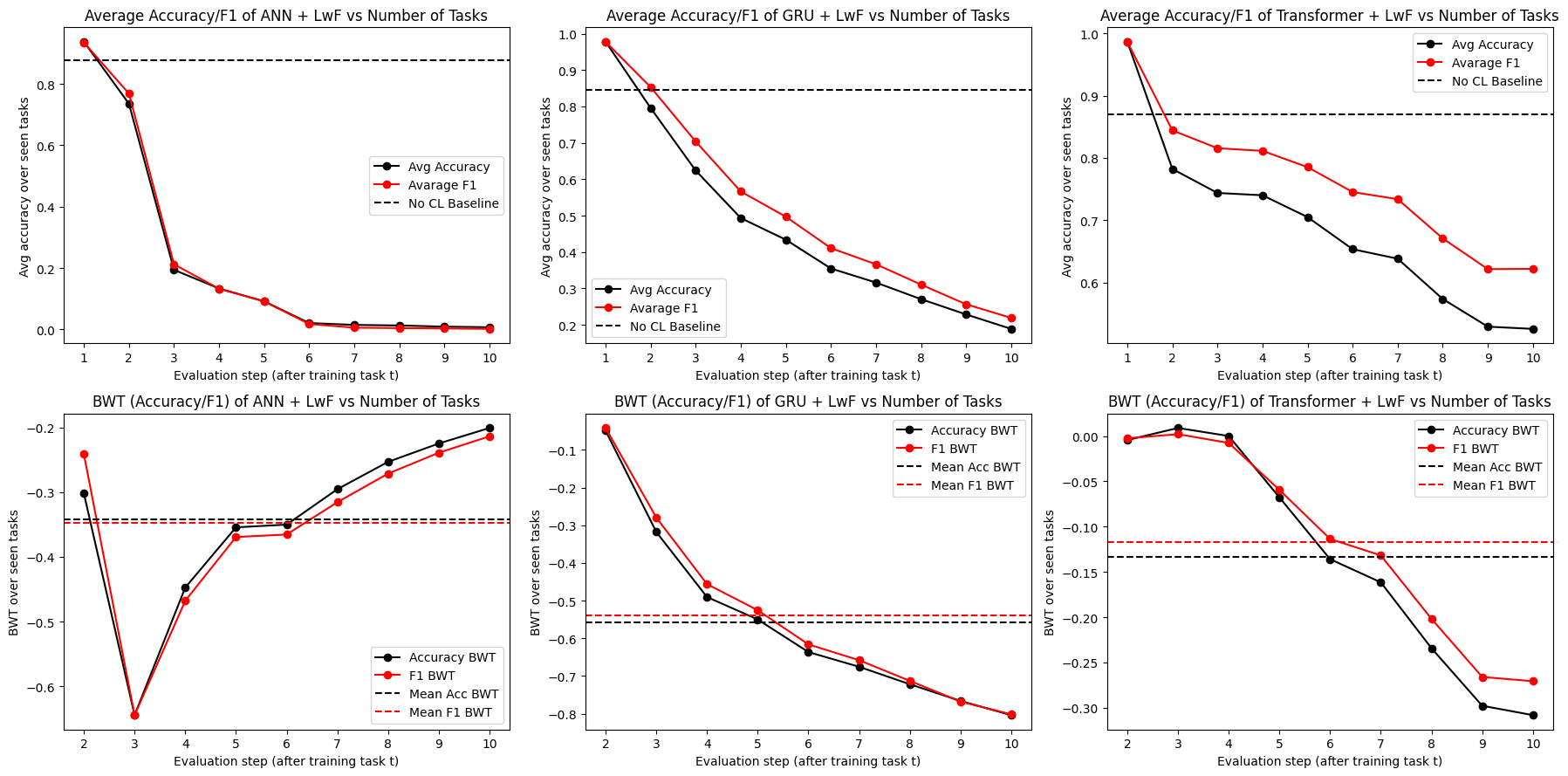}
  \captionof{figure}{AA, AF1, and average BWT over accuracy and F1 as they change over tasks for sequential training with LwF across all architectures.}
  \label{fig:lwf_plot}
\end{center}

\begin{center}
  \includegraphics[width=\textwidth]{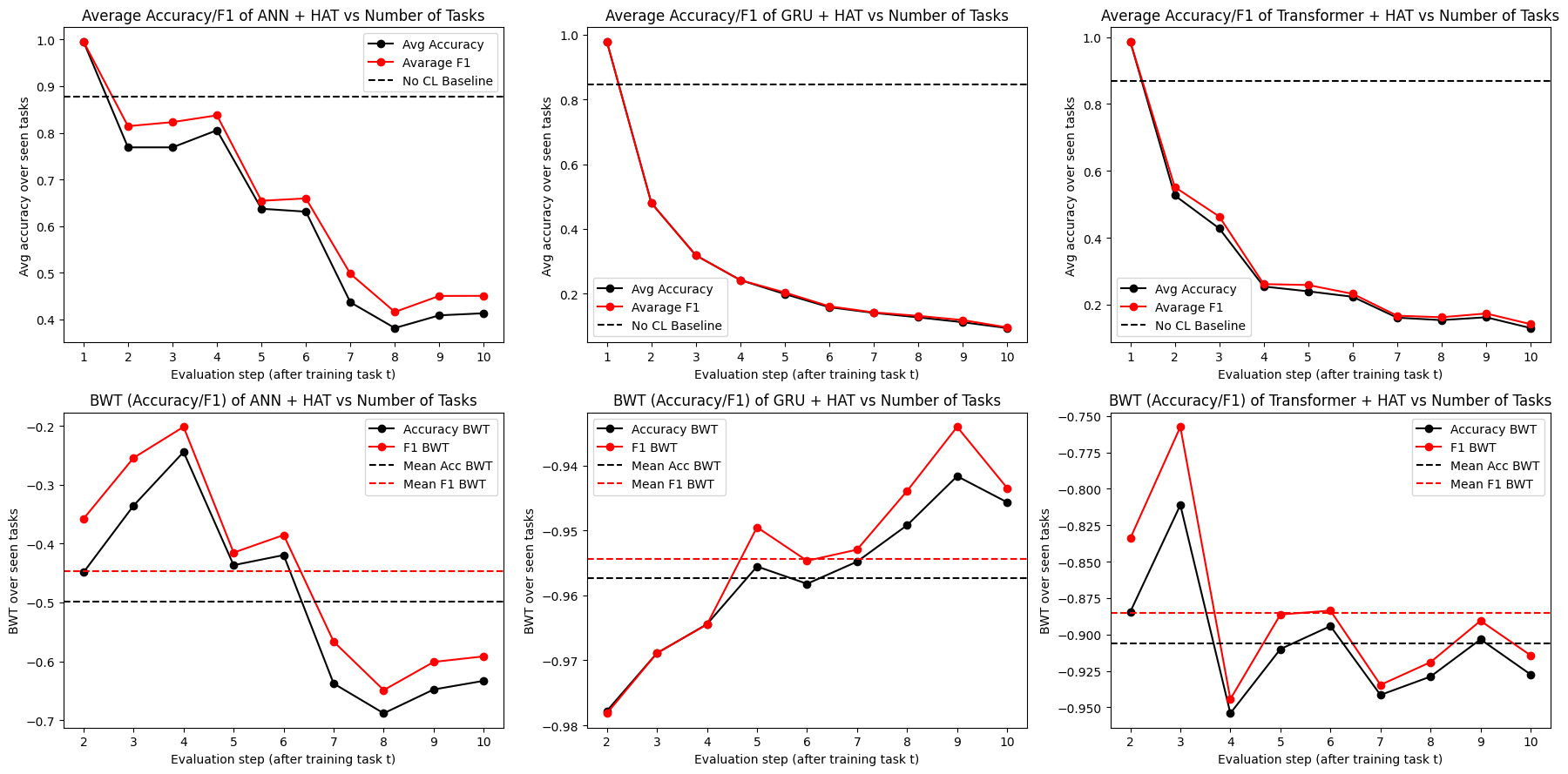}
  \captionof{figure}{AA, AF1, and average BWT over accuracy and F1 as they change over tasks for sequential training with HAT across all architectures.}
  \label{fig:hat_plot}
\end{center}
Afterwards, we evaluate all pairwise combinations of the three continual learning methods (MIR+LwF, MIR+HAT, LwF+HAT) for each architecture.
\newpage
\begin{center}
  \captionof{table}{MIR, LwF, and HAT Results on CLINC150 Dataset}
  \label{tab:cup_results}
  \begin{tabular}{p{1.5cm}p{4.6cm}p{4.6cm}p{4.6cm}}
      \toprule
      \textbf{Model} & \textbf{MIR+LwF} & \textbf{MIR+HAT} & \textbf{HAT+LwF} \\
      \midrule
      Artificial Neural Network (ANN) &
      \makecell[l]{$AA = 0.0249,\enskip AF1=0.0234,$\\ $\overline{BWT_{acc}} = -0.1486$\\ $\overline{BWT_{F1}} = -0.1719$} &
      \makecell[l]{$AA = 0.9738,\enskip AF1=0.9737,$\\ $\overline{BWT_{acc}} = 0.0002$\\ $\overline{BWT_{F1}} = 0.0001$} &
      \makecell[l]{$AA = 0.0502,\enskip AF1=0.053,$\\ $\overline{BWT_{acc}} = -0.2484$\\ $\overline{BWT_{F1}} = -0.3160$}\\
      \midrule
      Gated Recurrent Unit (GRU) &
      \makecell[l]{$AA = 0.732,\enskip AF1=0.8173,$\\ $\overline{BWT_{acc}} = 0.0005$\\ $\overline{BWT_{F1}} = 0.0003$}
      & \makecell[l]{$AA = 0.876,\enskip AF1=0.8657,$\\ $\overline{BWT_{acc}} = -0.0412$\\ $\overline{BWT_{F1}} = -0.0440$}
      & \makecell[l]{$AA = 0.2156,\enskip AF1=0.2638,$\\ $\overline{BWT_{acc}} = -0.5354$\\ $\overline{BWT_{F1}} = -0.5428$}\\
      \midrule
      Transformer Encoder
      & \makecell[l]{$AA = 0.7533,\enskip AF1=0.8313,$\\ $\overline{BWT_{acc}} = 0.0211$\\ $\overline{BWT_{F1}} = 0.0115$}
      & \makecell[l]{$AA = 0.8733,\enskip AF1=0.8696,$\\ $\overline{BWT_{acc}} = -0.0530$\\ $\overline{BWT_{F1}} = -0.0568$}
      & \makecell[l]{$AA = 0.324,\enskip AF1=0.3644,$\\ $\overline{BWT_{acc}} = -0.4577$\\ $\overline{BWT_{F1}} = -0.4118$}\\
      \bottomrule
  \end{tabular}
\end{center}

For ANN, only the MIR+HAT combination is truly effective: it reaches almost perfect performance with $AA=0.9738$ and $AF1=0.9737$, and essentially zero backward transfer ($BWT_{acc}\approx 0$,  $BWT_{F1}\approx 0$), meaning it both learns all tasks well and completely avoids forgetting.
In contrast, MIR+LwF and HAT+LwF remain close to chance level and still exhibit noticeable negative BWT, showing that for a simple feed-forward model, HAT only becomes useful when paired with replay.

For GRU and Transformer, combined methods again clearly outperform single ones.
On the GRU, MIR+HAT yields the highest accuracy ($AA=0.876$, $AF1=0.8657$) with only mild forgetting, while MIR+LwF also performs strongly ($AA=0.732$, $AF1=0.8173$) and even has slightly positive average BWT, indicating small beneficial backward transfer.
HAT+LwF, however, is much weaker and shows strong negative BWT.

A similar pattern appears for the Transformer: MIR+HAT achieves the best overall accuracy and F1 ($AA=0.8733$, $AF1=0.8696$) with small negative BWT, whereas MIR+LwF attains slightly lower AA/F1 but exhibits positive BWT, suggesting that combining replay with distillation can even improve earlier tasks.
HAT+LwF again underperforms.
Overall, combinations that include MIR (MIR+HAT, MIR+LwF) consistently deliver high final performance and near-zero (or slightly positive) BWT, showing that replay is the key ingredient and that regularization or parameter isolation work best as complements rather than standalone solutions.

\begin{center}
  \includegraphics[width=\textwidth]{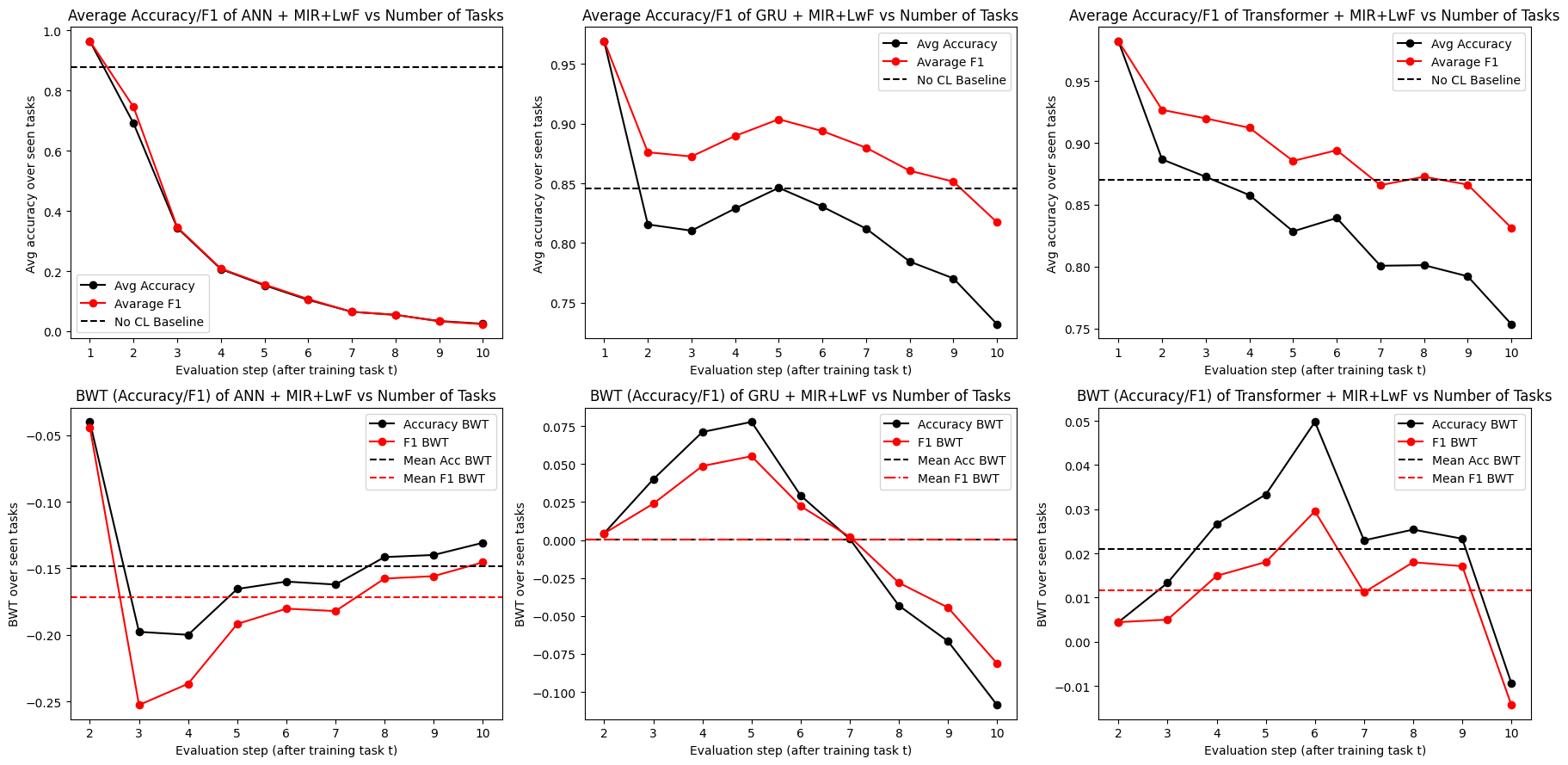}
  \captionof{figure}{AA, AF1, and average BWT over accuracy and F1 as they change over tasks for sequential training with MIR+LWF across all architectures.}
  \label{fig:mir+lwf_plot}
\end{center}

\begin{center}
  \includegraphics[width=\textwidth]{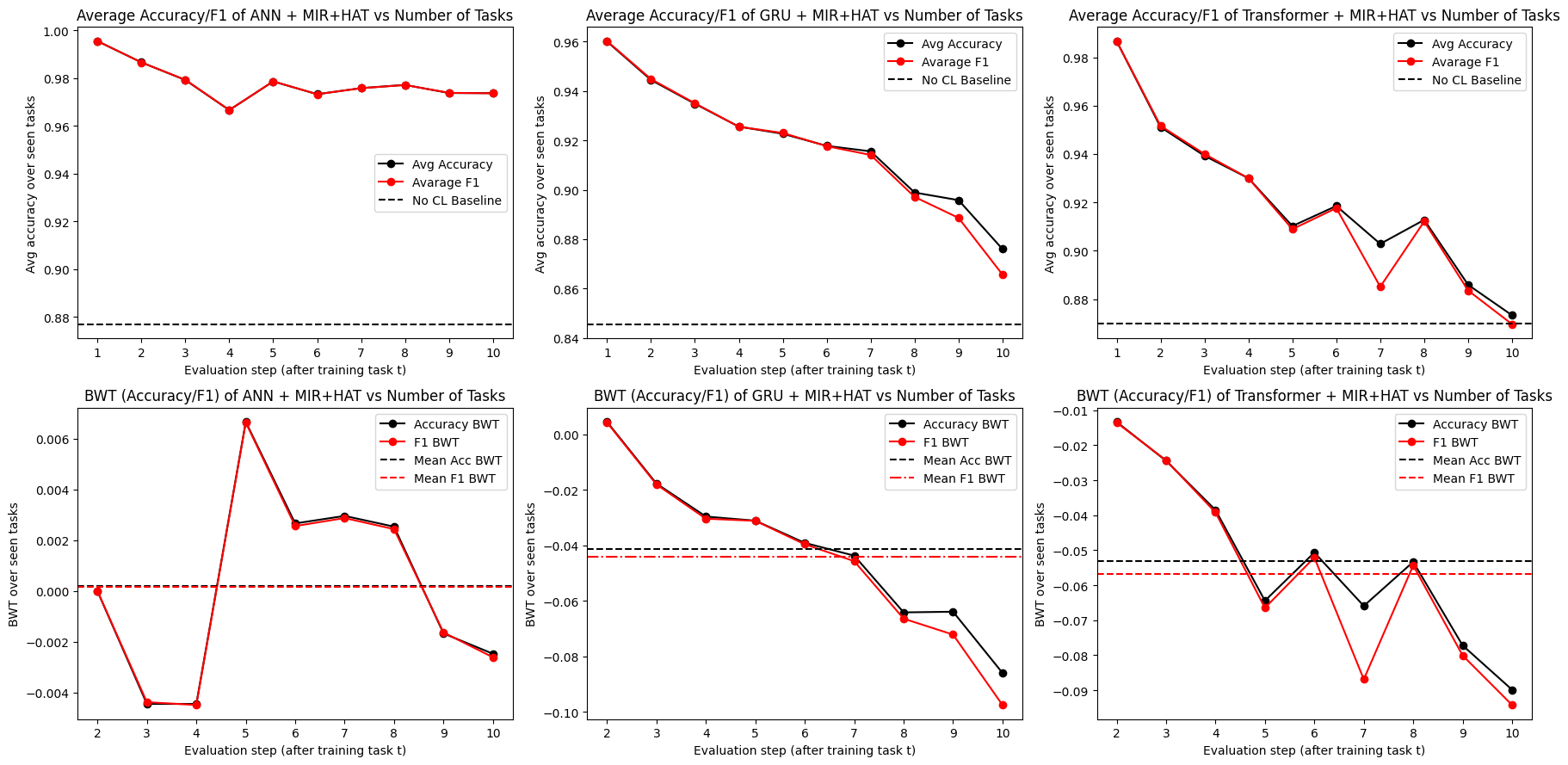}
  \captionof{figure}{AA, AF1, and average BWT over accuracy and F1 as they change over tasks for sequential training with MIR+HAT across all architectures.}
  \label{fig:mir+hat_plot}
\end{center}

\begin{center}
  \includegraphics[width=\textwidth]{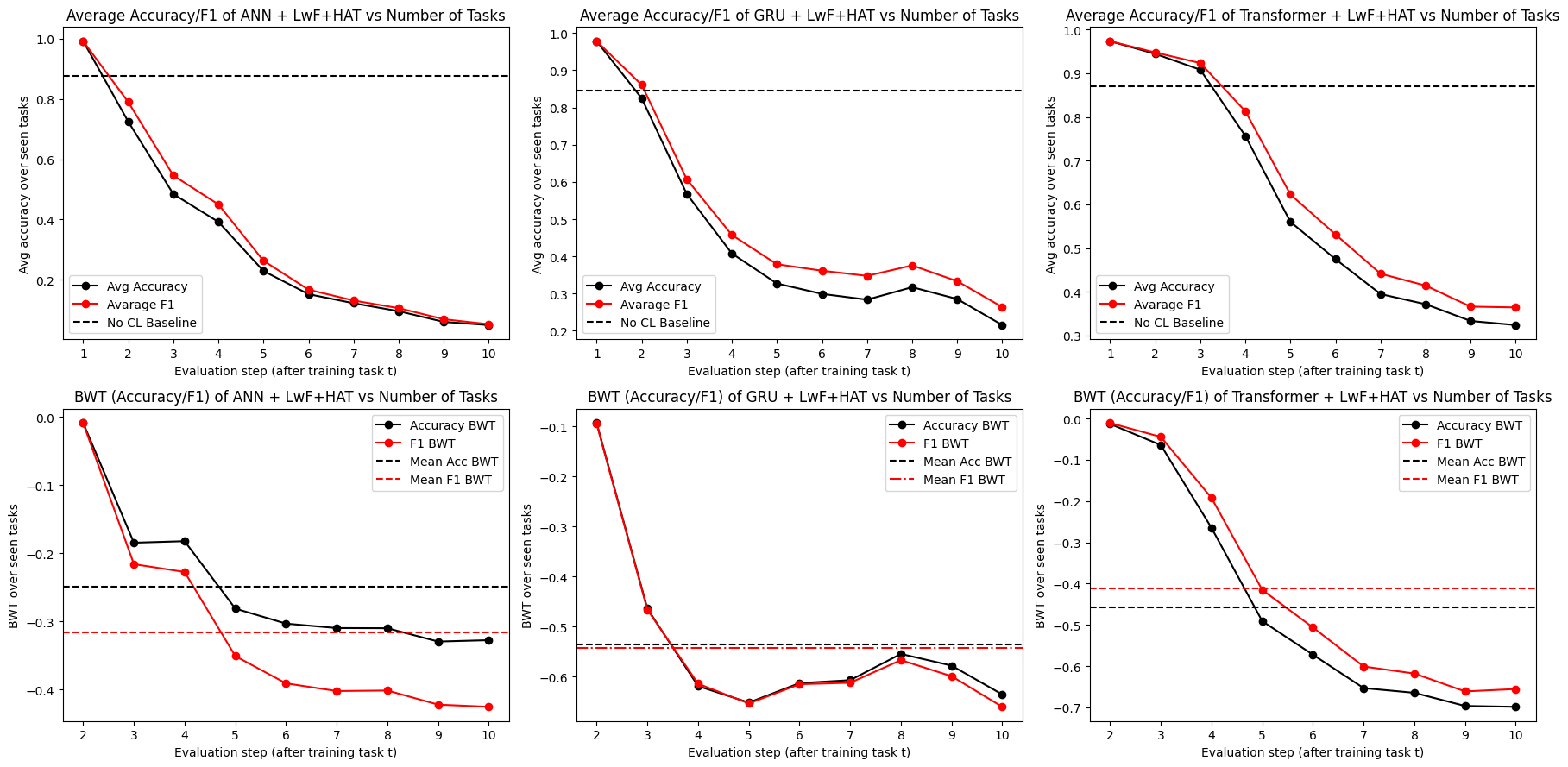}
  \captionof{figure}{AA, AF1, and average BWT over accuracy and F1 as they change over tasks for sequential training with LwF+HAT across all architectures.}
  \label{fig:lwf+hat_plot}
\end{center}
Lastly, we evaluate the unified combination of all three continual learning methods (MIR+LwF+HAT) for each architecture.
\begin{center}
  \captionof{table}{MIR+LwF+HAT Results on CLINC150 Dataset}
  \label{tab:all_results}
  \begin{tabular}{p{1.5cm}p{4.6cm}}
      \toprule
      \textbf{Model} & \textbf{MIR+LwF+HAT} \\
      \midrule
      Artificial Neural Network (ANN) &
      \makecell[l]{$AA = 0.8476,\enskip AF1=0.5956,$\\ $\overline{BWT_{acc}} = -0.0069$\\ $\overline{BWT_{F1}} = 0.0078$}\\
      \midrule
      Gated Recurrent Unit (GRU) &
      \makecell[l]{$AA = 0.9062,\enskip AF1=0.8781,$\\ $\overline{BWT_{acc}} = -0.0345$\\ $\overline{BWT_{F1}} = -0.0318$}\\
      \midrule
      Transformer Encoder
      & \makecell[l]{$AA = 0.8711,\enskip AF1=0.751,$\\ $\overline{BWT_{acc}} = -0.0225$\\ $\overline{BWT_{F1}} = 0.0349$}\\
      \bottomrule
  \end{tabular}
\end{center}

For the ANN, the triple method reaches $AA=0.85$, $AF1 = 0.6956$ with almost zero average backward transfer (slightly negative in accuracy and slightly positive in F1), indicating that it largely prevents forgetting while achieving strong, but not maximal, performance.
However, this is still worse than the best ANN configuration (MIR+HAT), which achieved almost perfect AA and AF1, suggesting that adding LwF on top of MIR+HAT can actually hurt the simple feed-forward model.

For the GRU, the unified MIR+LwF+HAT configuration achieves the best overall results across all GRU setups, with $AA=0.91$ and $AF1=0.88$, and only mild negative backward transfer.
This shows a clear synergy: replay, distillation, and parameter isolation complement each other particularly well for recurrent architectures.

For the Transformer, MIR+LwF+HAT attains accuracy comparable to MIR+HAT ($AA =0.87$), but with lower macro F1 and near-zero or slightly positive BWT, meaning it slightly improves stability at the cost of some class-wise performance.
Overall, the unified method is most beneficial for the GRU, offers a stability-focused trade-off for the Transformer, and is unnecessary or even slightly detrimental compared to MIR+HAT for the ANN.

\begin{center}
  \includegraphics[width=\textwidth]{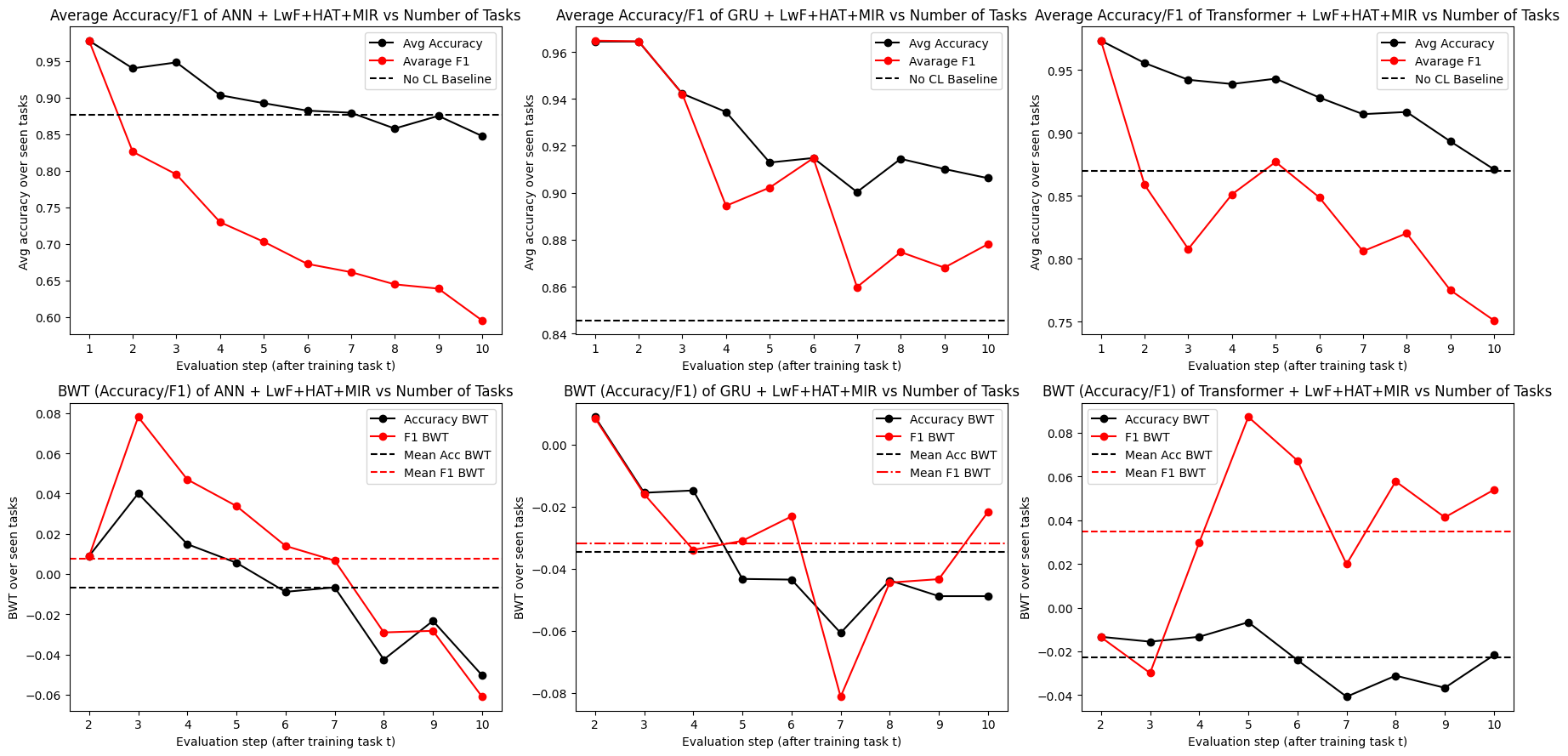}
  \captionof{figure}{AA, AF1, and average BWT over accuracy and F1 as they change over tasks for sequential training with MIR+LwF+HAT across all architectures.}
  \label{fig:mir+lwf+hat_plot}
\end{center}

\begin{center}
  \captionof{table}{Best continual learning configuration per architecture on CLINC150.}
  \label{tab:best_overall}
  \begin{tabular}{p{3cm}p{3.5cm}c c c c}
    \toprule
    \textbf{Model} & \textbf{Best CL approach} & \textbf{AA} & \textbf{AF1} & $\boldsymbol{\overline{BWT_{acc}}}$ & \textbf{Runtime} \\
    \midrule
    Artificial Neural Network (ANN) &
    MIR+HAT &
    0.9738 &
    0.9737 &
    0.0002 &
    Fastets \\
    \midrule
    Gated Recurrent Unit (GRU) &
    MIR+LwF+HAT &
    0.9062 &
    0.8781 &
    -0.0345 &
    Slowest\\
    \midrule
    Transformer Encoder &
    MIR+HAT &
    0.8733 &
    0.8696 &
    -0.0530 &
    Moderate\\
    \bottomrule
  \end{tabular}
\end{center}

\section{Discussion}
In this section we discuss the main empirical findings and relate them to the research questions posed earlier.
Overall, the resluts confirm that naive sequential fine-tuning is not sufficient for continual intent classification (Table \ref{tab:baseline_results}), but also show that there is no universal good choice for continual learning approach.
The effectiveness of a each strategy depends strongly on both the underlying neural network architecture and on whether methods are used individually or in combination.

When used in isolation (Table~\ref{tab:indiv_results}), replay-based MIR is consistently the strongest method across architectures, but it still exhibits substantial forgetting, especially for the simpler models.
For the ANN, MIR alone yields low final performance (\(AA \approx 0.19\), \(AF1 \approx 0.23\)) and strongly negative backward transfer.
LwF alone performs even worse (near-zero accuracy and F1 for the ANN), and HAT alone fails to stabilize learning for GRU and Transformer models (e.g., for the GRU with HAT we observe \(AA \approx 0.09\) and \(\overline{BWT_{acc}} \approx -0.96\)).
These results suggest that, in this setup, no single CL mechanism is sufficient: replay, distillation, and masking each address different aspects of catastrophic forgetting.

However, these results give an intuition that they can give better results when combined.
The combined methods show a dramatic improvement (Table~\ref{tab:cup_results}).
For the ANN, MIR+HAT almost completely solves catastrophic forgetting, achieving \(AA \approx 0.97\), \(AF1 \approx 0.97\), and essentially zero average backward transfer.
For the GRU and Transformer, combinations that include MIR again dominate: MIR+HAT and MIR+LwF both deliver high final performance, while HAT+LwF remains comparatively weak.
Importantly, some combinations (e.g., MIR+LwF for GRU and Transformer) achieve slightly positive average backward transfer, indicating that revisiting carefully selected past examples with distillation can even improve performance on earlier tasks.

The unified configuration MIR+LwF+HAT (Table~\ref{tab:all_results}) further clarifies the picture.
For the GRU, this triple combination yields the best overall performance across all GRU setups (\(AA \approx 0.91\), \(AF1 \approx 0.88\)) with only mild negative backward transfer, suggesting a genuine synergy between replay, regularization, and parameter isolation.
For the Transformer, the triple method reaches accuracy comparable to MIR+HAT but with slightly lower macro F1 and near-zero or slightly positive backward transfer, pointing to a trade-off between class-wise performance and stability.
For the ANN, however, MIR+HAT remains the best configuration; adding LwF on top of MIR+HAT lowers performance while offering little additional stability benefit.
This asymmetry highlights that the same CL components interact differently depending on the underlying architecture.

The final best results of each architecture with the respective best CL approach yield better results than by the joint training, meaning that continual learning methods can help generalization by acting as a regularizer.

On of the central goals of this work was to balance stability and plasticity.
In our results we observe that inidividual methods often favour one side of the trade-off.
For instance, MIR alone improves plasticity by revisiting past samples, but still allows some forgetting, especially for simpler models.
LwF alone tends to prioritize stability, but at the cost of plasticity, leading to poor learning of new tasks.
HAT alone can strongly protect past knowledge, but may overly restrict plasticity, especially for complex architectures like GRU and Transformer.
However, when combined, these approaches tend to balance each other out, leading to both high final performance and near-zero or slightly positive backward transfer.
MIR+HAT for the ANN achieves both high final performance and almost zero backward transfer, demonstrating that replay can supply the necessary plasticity while HAT protects previously used capacity.
For GRU and Transformer, the combinations MIR+LwF and MIR+LwF+HAT achieve high accuracy and F1 with near-zero or mildly negative backward transfer, indicating that the network can integrate new tasks without catastrophically overwriting old ones.
In several cases, slightly positive BWT values suggest beneficial backward transfer, where learning new tasks indirectly helps refine earlier decision boundaries.
The overall picture suggests that replaying past examples with MIR is the most crucial ingredient for plasticity, while LwF and HAT act as complementary mechanisms and help stabilize learning and prevent forgetting, leading to a more balanced stability--plasticity trade-off.
Using these methods in isolation either leaves too much forgetting or overly restricts learning, whereas carefully chosen combinations can approximate the performance of joint training while respecting the sequential constraint.

The three architectures exhibit distinct behaviours under continual learning.
In general, the Transformer achieves the highest absolute performance when equipped with strong CL methods (e.g., MIR+HAT), reflecting its greater capacity and ability to model long-range dependencies.
However, the relative gains from CL are largest for the simpler ANN.
Moving from individual methods to MIR+HAT transforms the ANN from a highly forgetful model into one that yields a superior result on joint-training.
This suggests that even lightweight architectures can become competitive continual learners when equipped with an appropriate combination of replay and parameter isolation.

The GRU occupies an intermediate regime.
Without CL, recurrent models already capture useful sequential structure but still suffer severe forgetting.
With MIR+LwF+HAT, the GRU attains both high final accuracy and relatively stable backward transfer, indicating that recurrent architectures can particularly benefit from the combined effects of replay, distillation, and masking.

Across all architectures, the qualitative ranking of CL strategies is broadly consistent: replay-based methods and hybrids that include MIR clearly outperform naïve sequential fine-tuning and individual LwF or HAT runs.
At the same time, the best-performing combination differs slightly by architecture (MIR+HAT for ANN and Transformer, MIR+LwF+HAT for GRU), showing that the interaction between CL mechanisms and model inductive biases is non-trivial.
Practically, this implies that CL method selection should consider not only the task and data but also the backbone architecture.

Furthermore, the GRU and Transformers are data-hungry models and may require more data or pretraining to fully leverage their capacity.
\newpage
\section{Limitations}
From a practical perspective, our findings suggest several guidelines for designing continual intent classification systems.
First, replay should be considered a default choice.
Methods without a replay component either perform poorly or require strong architectural constraints.
Second, adding either distillation (LwF) or parameter isolation (HAT) on top of replay can significantly improve stability, but the best combination is architecture-dependent.
The experiments demonstrate that continual learning for NLP is feasible with carefully designed combinations of replay, regularization, and parameter isolation, and that understanding the interaction between CL mechanisms and model architecture is crucial for achieving a favourable stability--plasticity trade-off.
At the same time, several limitations of this study should be taken into account and acknowledged when interpreting the results.
First of all, we only evaluated on a single dataset (CLINC150) and a specific task (intent classification) with a fixed 10-task label-disjoint split.
Although this setup is useful for controlled analysis, it may not fully capture the diversity of real-world continual learning scenarios, such as domain shifts, evolving label spaces with overlaps, or truly open-ended streams of user queries.
Future work therefore should assess the generality of these findings across multiple datasets (e.g., other intent benchmarks, sentiment analysis, or topic classification), tasks, including partially overlapping label sets and domain-incremental settings.
Secondly, our analysis is limited to three backbone architectures: a feed-forward ANN, a GRU, and a lightweight Transformer encoder.
While these models represent common text-classification paradigms, they do not cover the full spectrum of modern NLP architectures, particularly large pretrained language models and parameter-efficient fine-tuning techniques.
Investigating how MIR, LwF, HAT, and their combinations interact with large-scale Transformers, adapters, or prompt-tuning methods is an important direction for future work, especially in resource-constrained continual learning settings.
Finally, although we tune hyperparameters using Bayesian optimization, our study does not explore the sensitivity of each method to memory size, replay frequency, regularization strength, or mask sparsity.
We also do not perform formal statistical significance testing across random seeds.
A more detailed study of hyperparameters, memory size, and computational cost would make our conclusions more reliable and give clearer guidelines for using continual learning in real-world intent classification systems.

\section{Conclusion}
In this work, we studied catastrophic forgetting in a continual intent classification setting based on the CLINC150 dataset.
We constructed a 10-task label-disjoint scenario and compared three neural architectures: a feed-forward ANN, a GRU, and a Transformer encoder under three continual learning families: replay (MIR), regularization (LwF), and parameter isolation (HAT), both individually and in combination.
Using accuracy, macro F1, and backward transfer as evaluation metrics, we analyzed how different strategies balance stability (retaining old tasks) and plasticity (learning new ones).

Our results show that naive sequential fine-tuning suffers from severe forgetting across all architectures, while no single CL method completely solves the problem.
Replay-based MIR is the most reliable individual strategy, but it still leaves noticeable negative backward transfer, especially for simpler models.
The strongest performance is obtained by combining methods: MIR+HAT almost eliminates forgetting and approaches joint-training performance for the ANN and Transformer, whereas the triple combination MIR+LwF+HAT yields the best overall results for the GRU.
These findings highlight that replay is a crucial component for plasticity, while regularization and parameter isolation act as complementary mechanisms that improve stability when used alongside replay.

Architectural differences also play a clear role.
The Transformer achieves the highest absolute performance with strong CL methods, but the relative gains from CL are largest for the ANN, which transforms from highly forgetful to almost perfectly stable when equipped with MIR+HAT.
The GRU benefits most from the full MIR+LwF+HAT combination, indicating that recurrent models can particularly exploit the synergy between replay, distillation, and masking.
Overall, the qualitative ranking of CL strategies is consistent across architectures, but the best-performing combination is architecture-dependent.

Taken together, our experiments suggest that effective continual learning for intent classification requires both an appropriate backbone and a carefully chosen combination of replay, regularization, and parameter isolation.
We hope that this analysis, and the concrete configurations identified as strong baselines, can serve as a reference point for future work on continual learning in NLP and for practitioners interested in deploying continual intent classifiers in real-world systems.












\newpage
\bibliographystyle{unsrtnat}
 
\end{document}